\documentclass{article}
\usepackage{cec2003,multicol,times}

\usepackage{epsfig}

\usepackage{amssymb}

\usepackage{amsmath}

\begin{document}
\pagestyle{empty}
\sloppy

\twocolumn[
\title{An efficient memetic, permutation-based evolutionary algorithm for real-world train timetabling}
%
%
\vspace{0.1in}
\begin{multicols}{2}
\begin{center}
\textbf{Yann Semet}  \\
Equipe TAO, INRIA Futurs \\
Orsay, FRANCE \\
semet@lri.fr
\end{center}
\begin{center}
\textbf{Marc Schoenauer}  \\
Equipe TAO, INRIA Futurs \\
Orsay, FRANCE \\
marc@lri.fr
\end{center}
\end{multicols}
\vspace{0.25in}
] 

\begin{abstract}
Train timetabling is a difficult and very tightly constrained combinatorial problem that deals with the construction of train schedules. We focus on the particular problem of local reconstruction of the schedule following a small perturbation, seeking minimisation of the total accumulated delay by adapting times of departure and arrival for each train and allocation of resources (tracks, routing nodes, etc.).

We describe a permutation-based evolutionary algorithm that relies on a semi-greedy heuristic to gradually reconstruct the schedule by inserting trains one after the other following the permutation. This algorithm can be hybridised with ILOG commercial MIP programming tool CPLEX in a coarse-grained manner: the evolutionary part is used to quickly obtain a good but suboptimal solution and this intermediate solution is refined using CPLEX. Experimental results are presented on a large real-world case involving more than one million variables and 2 million constraints. Results are surprisingly good as the evolutionary algorithm, alone or hybridised, produces excellent solutions much faster than CPLEX alone. 
\end{abstract}

\section{Context and Rationale}

\subsection{Context}

Similarly to numerous complex systems met in organisations or
operations research contexts, 
suburban transportation networks feel a growing need for automated
computational optimisation in a wide variety of areas. As these networks grew in complexity and size, such a need has
emerged from the joint pressure of marketing strategies, cost
reduction policies and environmental concerns.

Among these needs, improving punctuality stands out as an important
factor of consumer satisfaction and plays a key role in the railway
world race for a competitive edge. The problem we are dealing with
here is concerned with diminution of resulting delays in case of small
perturbations of the traffic. More precisely, when for some
operational reason, a train is delayed for a few minutes at some
point of the network (unavailable track, equipment malfunctioning,
etc.), it involves taking the right decisions for space-wise and
time-wise neighbouring trains so that the consequence is minimal. In
more mathematical terms, the problem is to locally reconstruct the
schedule, by wisely delaying times of departure and arrival as well as
modifying track allocation. The objective function is the difference,
in terms of delay, between the original schedule and the recomputed
one. This reconstruction has to be done while enforcing a great deal
of operational, commercial and safety constraints that strongly limit
the possibilities. The problem at hand is therefore a difficult,
combinatorial, constrained optimisation problem, equivalent to some form of job-shop scheduling problem.

At the moment, this problem that empirically appears as being NP-hard, is
practically solved in real time by human operators who make use of
expert knowledge and common sense to perform optimisation. A first
scientific investigation of the problem has led to its description and
solving as a Mixed Integer Program (MIP) with ILOG commercial tool 
CPLEX. Obtained results are satisfactory for they show the adequacy of
the chosen model and prove its computational tractability but do not
meet criteria for real-time exploitation and suggest weakness with
respect to combinatorial explosion. They indeed show that several
hours are necessary on a fast computer to reach acceptable solutions
for average-size instances of the problem. Additionally, computational
efficiency seems to be greatly affected by growth of the instance in
either size or complexity, which eventually leads to
intractability. 

\subsection{State of the Art and Chosen Approach}

Modern stochastic optimisation techniques, and especially Evolutionary
Algorithms are specifically tailored to handle such situations. They
can efficiently browse large, irregular search spaces and quickly
provide their practitioners with good solutions without requiring
deep, explicit knowledge of the problem. They additionally offer a wide
variety of techniques to handle constraints
\cite{Michalewicz96a}. 

An early example of applying GAs to scheduling problems is Davis's work in 1985 \cite{DavisScheduling}. From then on, scheduling has been a great field of application for evolutionary algorithms. One can refer to \cite{FangScheduling,husbandsScheduling,specialIssue,Globus03Scheduling} for surveys and a few recent examples. A number of teams worked on the particular case of train scheduling. We give a list, by no means exhaustive, of important examples. Caprara et al. \cite{caprara00modeling,caprara01solution} offered a MIP formulation of the problem along with a solving algorithm which make use of graph-theoretic techniques and Lagrangian relaxation. Parkes and Ungar \cite{ParkesTrain} use market algorithms where trains are represented by virtual agents who competitively interact for the allocation of resources attributed by an efficient auction-based system. Kwan et al. propose a Co-evolutionary approach \cite{kwan:techReport,kwan:CEC03,kwan:GECCO03} for initial timetable generation. Finally, Juill\'e \cite{juille04trains} mixes permutation-based evolutionary search and constraint oriented programming to solve a bi-objective instance of the train timetabling problem in a decision support context. 

Following Juill\'e, our approach is an indirect one : the genotype is an
ordering of the trains (a permutation), but all constraints
are handled in some scheduler, i.e. during the morphogenesis process
transforming the permutation into a valid schedule.

This paper starts by precisely describing the problem, including
constraints, degrees of freedom and objective function, in Section
\ref{problem}.  A global picture of the implemented system is then
given in section \ref{global}, illustrated by a flowchart. The two
following sections give respectively technical details on scheduling
heuristics (section \ref{scheduler}) and evolutionary technical choices
(representation, operators, etc.) -- Section \ref{evolutionary}. The
final section offers evidence of the efficiency of the chosen approach
compared to the purely deterministic method using CPLEX alone on a
real-world instance of the problem and draws appropriate qualitative
conclusions.

\section{The problem}
\label{problem}
This section gives details on the mathematical model underlying the
scheduling problem, namely the objective function, the degrees of
optimisation freedom and enforced constraints. It also introduces
symbols, notations and elementary assumptions. 

\subsection{Degrees of freedom}

\begin{figure*}[htbp]
    \begin{center}
  \epsfig{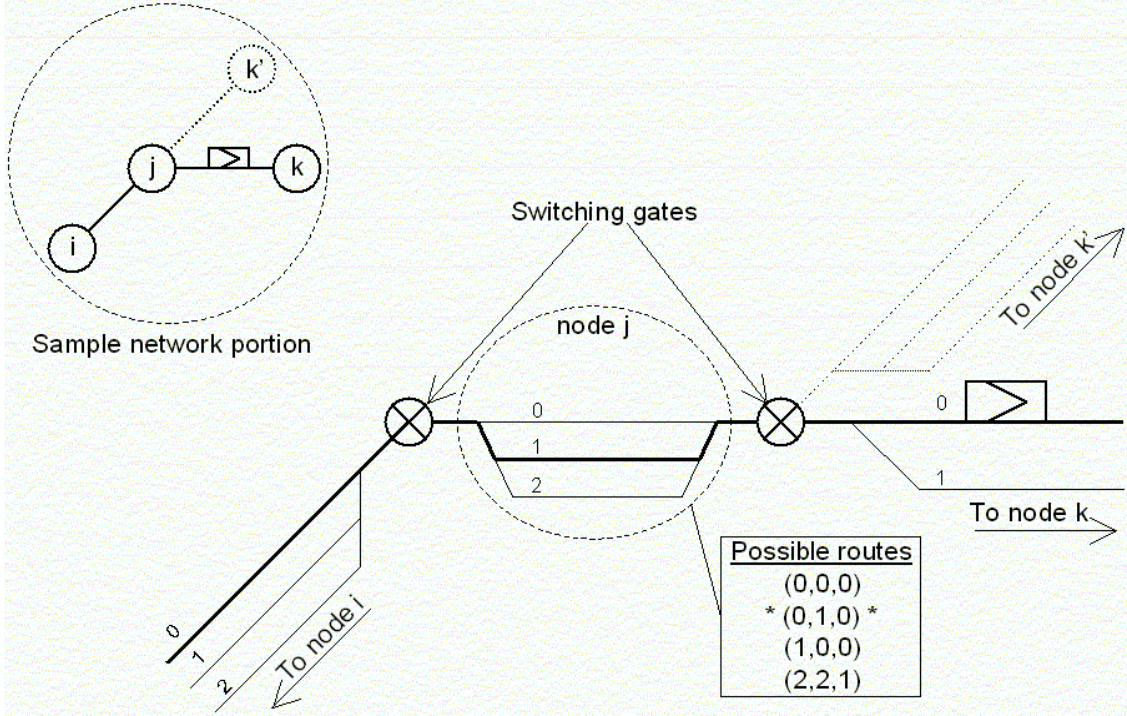}      
      \caption{A sample network portion}
      \label{fig:samplePortion}
    \end{center}
~\\ 
\end{figure*}

As illustrated by figure \ref{fig:samplePortion}, the railway network can be seen as a graph where nodes are stations or
switchings and where interconnecting edges eventually hold several
tracks for trains to use. Each train $c$ ($C$ being the set of all
trains \footnote{We choose $c$ instead of $t$ to avoid confusion with
time.}) has a fixed ordered list of nodes to visit $I(c)$. 
There
are three degrees of freedom for each train at each at node : times of
arrival $a$, departure $d$ and track choice $r$ (for
\emph{route}). $a$ and $d$ are integers giving the number of seconds
elapsed since an arbitrary temporal origin. Track choice actually
implies three tracks to be chosen: one in both incoming and outgoing
edges (resp. $u_{inc}$ and $u_{out}$) and one inside the node
($u$). These three decisions are linked by underlying physical
constraints: picking, for instance, a particular incoming track
restricts the number of possible subsequent tracks inside the node and
in the outgoing edge. Each node therefore holds a list of triplets
(i,j,k) indicating the physically possible combinations among which a
choice has to be made. To sum up, a schedule is completely defined by assigning values to all degrees of
freedom, i.e., for each train $c$ at each node $i$

\begin{eqnarray}\label{eq:dof}
d.o.f.(c,i)=
\left\lbrace
\begin{array}{l}
a(c,i) \in \mathbb{N}\\
d(c,i) \in \mathbb{N}\\
r(c,i)=(u_{inc},u,u_{out}) \in \mathbb{N}^3
\end{array}
\right.
\end{eqnarray}

The problem at hand is to reconstruct a valid schedule after some
incident. The initial schedule that all trains had been assigned
before the incident  will be referred to by zero-indices
(e.g. $a_0(c,i), \ldots$).

\subsection{Visualisation}

\begin{figure}[htbp]
    \begin{center}
        \epsfig{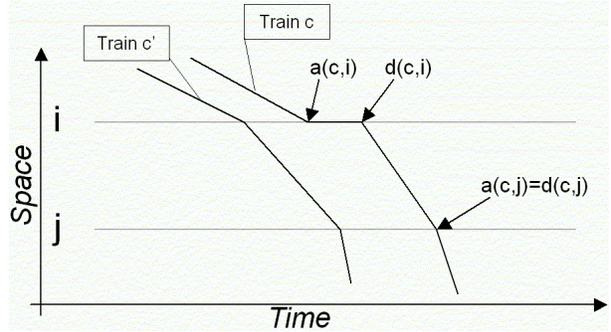}      
      \caption{A space/time diagram}
      \label{fig:spaceTime}
    \end{center}
\end{figure}

As illustrated by self-explanatory Figure \ref{fig:spaceTime},
schedules can be represented using space/time diagrams. This visual
representation is very useful to railway scheduling experts to get a
global glimpse of a complex network and to quickly detect
abnormalities. The reader might also find it useful to think the
equations given below in those visual terms and to see it as a visual
representation of the phenotype (see Section \ref{global}). 

\subsection{Constraints}

This sections enumerates the constraints limiting the aforementioned degrees of freedom. All of them are hard constraints and cannot possibly be violated, mostly for safety reasons. This set of constraints is not exactly exhaustive with respect to real world operational conditions but gets pretty close, which definitely places the problem outside the ``toy problem'' category. 

For clarity, quantifiers (e.g. $\forall c$, $\forall i$, etc.) are omitted in all equations given below, assuming inequalities apply to the obvious relevant sets. For instance, safety spacing constraints only apply between trains that use the same resource (node or track section) at some point.

All Greek letters, introduced in alphabetical order, will represent prescribed constants for spacing intervals.

\subsubsection{Initial times}

These constraints are the most obvious ones: trains cannot arrive or leave a node earlier than specified in the initial schedule.
\begin{eqnarray}
        \left\lbrace
        \begin{array}{l}
        a(c,i)\ge a_0(c,i)\\    
        d(c,i)\ge d_0(c,i)\\
        \end{array}
        \right.
\end{eqnarray}

\subsubsection{Stopping time}
For both maintenance and commercial reasons, trains stopping times are both upper and lower bounded: 
\begin{eqnarray}
        \alpha_{min}(c,i) \leq d(c,i)-a(c,i)\leq \alpha_{max}(c,i)
\end{eqnarray}

\subsubsection{Speed}

According to physical contingencies such as engine power, trains need a certain time to cover the distance between two nodes, say $i$ and $i'$:

\begin{eqnarray}
        a(c,i')-d(c,i) \geq \beta(c,i\to i') 
\end{eqnarray}

\subsubsection{Safety Spacing}

At all times, there must be sufficient distance between any pair of trains so that either has enough time to undertake and complete emergency braking procedures. This applies for edges:
\begin{eqnarray}
\left\lbrace
\begin{array}{l}
d(c',i)\geq d(c,i)+\gamma(c,c',i\to i')\\
a(c',i')\geq a(c,i')+\gamma(c,c',i\to i')\\
\end{array}
\right.
\end{eqnarray}
as well as for nodes:
\begin{eqnarray}
a(c',i)\geq d(c,i)+\gamma(c,c',i)
\end{eqnarray}

\subsubsection{Connections}

A number of trains are scheduled to go back and forth along a given path, usually between two large cities. For the sake of homogeneity, a train making several such round-trips over the course of a problem instance makespan will be appear in the data as several trains, one for each one-way trip even if there is only one physical train. This imposes a number of constraints to be enforced at extreme nodes where the connection between pseudo trains takes place. For instance, for a shuttle between $i'$ and $i$, if train $c$ went from $i'$ to $i$ and is about to go back (under the name $c'$), that is from $i$ to $i'$, one has to have:

\begin{eqnarray}
\left\lbrace
\begin{array}{l}
d(c',i)=d(c,i)\\
a(c',i')=d(c',i)+\delta_{min}(c',i)\\
r(c',i)=r(c,i)\\
\end{array}
\right.
\end{eqnarray}

\subsubsection{Switching gates}

Two edges connected to one same node can be sharing switching gates. These gates have limited capacities, being able, for example, to handle only one train at a time. Spacing constraints appear as a consequence to ensure proper use of this shared physical resource. These constraints are responsible for a very large part of the problem's combinatorial complexity. They are separated in four categories according to what kind of edges (incoming (I) or outgoing (O)) they apply to: II,IO,OI or OO. For instance:

\begin{eqnarray}a(c',i)\geq a(c,i)+\epsilon^{OI}(c,c',i)\end{eqnarray}

\subsection{Objective function}

The goal of the optimisation procedure is to minimise the total accumulated delay, i.e. for all trains at all nodes, the difference between the actual time of arrival and the theoretical one. Delays are caused by one small perturbation of the network. Small means that only one train is delayed for a few minutes at some point of the network (inside a node or in between two nodes) for operational reasons. Because of the highly constrained status of the problem, such a small perturbation is sufficient to yield a great deal of subsequent delaying that affect many surrounding trains. 

Using the above notations, the fitness function can be written as
\begin{eqnarray}\label{eq:foncObj1}
        f=\sum_{c \in C}{\sum_{i \in I(c)}{a(c,i)-a_0(c,i)}}
\end{eqnarray}

Because the theoretical schedule is fixed, one can therefore equivalently consider as fitness:

\begin{eqnarray}
        f=\sum_{c \in C}{\sum_{i \in I(c)}{a(c,i)}}
\end{eqnarray}

\section{Algorithm: the global picture}
\label{global}
This section gives an general outline of the final hybrid algorithm that is proposed to solve the problem described in previous section. The algorithm is an hybrid, or memetic algorithm, in the sense that it combines an evolutionary engine with a mathematical programming tool, namely ILOG CPLEX. This combination aims at getting the best of two worlds: the relative speed and efficiency of global stochastic search when facing combinatorial explosion with the rigour and exhaustivity of local exploitation.

As a first step, that is for the initial research described in this paper, we proceed in a ``coarse-grained'' fashion: the evolutionary part of the algorithm is used to quickly obtain a good but suboptimal solution. This solution, the best individual in the population after $K$ generations, is fed to CPLEX as an initial solution, a starting point for its search for the global optimum. Future efforts will be concerned by some ``finer-grained'' hybridisation, which would mean a more frequent exchange of solutions between the two modules: for instance, every 5 generations, the Evolutionary Algorithm could try to give an intermediate solution to CPLEX in order to see whether a satisfactory basin of attraction has been reached.

It is important to notice that both parts of the algorithm, the Evolutionary Algorithm and CPLEX, are independent and autonomous. It means both can solve the problem but either suboptimally (the EA, discussion below) or too slowly (CPLEX). The two following subsections give technical details first on how we adapt the evolutionary paradigm to this problem and then on how the hybridisation scheme works. 

\subsection{Coupling permutations and a scheduling heuristic}

Two fundamental kinds of approaches can be followed when solving a problem with evolutionary algorithms: direct and indirect ones. A direct approach straightforwardly encodes, generally using a specifically adapted representation, the degrees of freedom of the problem. The variation operators involves either standard operators (e.g. Gaussian mutation, multi-point crossover, etc) or specific ones (e.g. to ensure some constraints are preserved). An indirect approach uses an additional mapping function which is devoted to constructing a solution starting from a restricted number of parameters. In the latter case, the evolutionary algorithm browses a smaller, usually much better structured search space which is additionally richer in terms of information relevant to the problem. These advantages are strong but not always necessary however and in any case induce a large additional development cost for the mapping function, which besides demands expert knowledge and therefore deprives the evolutionary optimisation toolbox from its ``black-box'' quality. Nevertheless,  for some problems, typically scheduling problems like the one described above, the direct approach is impractical for it creates a huge, ill-structured search space but also and mostly because it offers no easy way to explicitly enforce constraints.

Thus we have chosen to use an indirect approach based on a permutation representation. This approach is classical for Job-Shop or Timetabling problems (see \cite{} for example). The mapping function in this case constructs the solution by iterative greedy insertion of elements (tasks, planes, etc. in our case trains) in the order dictated by the permutation. \emph{Greedy} means that when a train is placed in the schedule, it uses the available resources, as left by previously scheduled trains, in the optimal way. The underlying idea is that the evolutionary optimisation process should come up with the permutation representing the order in which trains should be allowed to use available resources (tracks, gates, \ldots) and the priority decreasingly according to which they could be forced to wait for fluidification purposes. The order resulting from the Evolutionary Algorithm should be, if not optimal, natural and efficient with respect to the perturbation that created the problem.

More precisely, we manipulate a permutation of trains. These trains
are iteratively inserted in an initially empty schedule. Each train is
inserted node by node, greedily (see Section \ref{scheduler}), 
which means that for each node in
turn, the degrees of freedom are set to the best possible
combination. As we are minimising total accumulated delays, this best
combination corresponds to the track choice and time settings which
allow for the earliest possible departure time. The difficulty of
course lies in doing so while respecting all of the constraints described in Section \ref{problem}. The
procedure in charge of finding this locally optimal setting for a train is
called the \emph{scheduler}. 

An important thing to notice along the way is that such scheduler is actually semi-greedy
because optimality is enforced only at the node level as opposed to
greedy \emph{stricto sensu}, which would mean optimal at the train
level. It might indeed be profitable for the train to be suboptimal at
some node to later catch up and be globally faster. But implementing a
scheduler permitting globally greedy insertions would have been a much
more complex task, and would have yielded a much slower algorithm.  The
bad consequence is that the best solutions found by the algorithm are
suboptimal. This matter will be further discussed later on.

The evolutionary algorithm thus evolves a population of
permutations. These permutations (the genotypes of the Evolutionary
Algorithm) are turned into proper schedules (the phenotypes) by the
scheduler. Such schedules are then 
evaluated for total accumulated delay, with respect to the initial,
unperturbed schedules, to provide fitness. This population goes
through the traditional evolutionary loop (see Section
\ref{evolutionary}) until it 
reaches a satisfactory solution. The algorithm is then either stopped or
continued, as described in next sub-section.

\subsection{Coarse-grained hybridisation}

CPLEX is a Mixed Integer Programming (MIP) tool developed and sold by the ILOG corporation. It solves linear programs, a category in which our problem, as described above, perfectly fits. It does so by combining repeated uses of the simplex method with a Branch and Bound algorithm that explores a real-numbered version of the integer problem. It also make use of a great deal of various heuristics and cuts to efficiently reduce or manipulate the problem. When used alone, it proceeds in two steps. First, it tries to come up with an initial, feasible solution using constructive heuristics. Then, it browses the feasible space following its Branch and Bound strategies. It can also accept a precomputed initial feasible solution as a starting point. That is how the coarse-grained hybridisation works: the GA, after a number of generations provides CPLEX with the solution, or schedule, produced by its best individual.

\begin{figure}[htbp]
    \begin{center}
        \epsfig{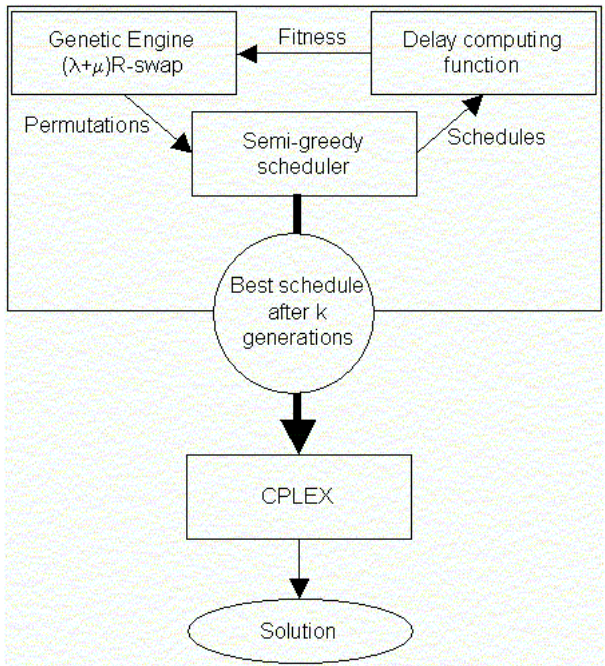}      
      \caption{The global picture}
      \label{fig:flowChart}
    \end{center}
\end{figure}

The relationships between the algorithm's various components are illustrated by the flowchart given in figure \ref{fig:flowChart}.

\section{A semi-greedy scheduler}
\label{scheduler}
The ``scheduler'' procedure is at the very heart of our algorithm, it manipulates its underpinnings and is the basis on which much of the resulting performance is going to come from. This section gives a sketchy outline of its functioning without delving in technicality.

The role of the scheduler is to read the permutation of trains and to place them, each in turn, in the schedule while respecting all of the specified constraints. The following sections explain how constraints are checked for and how conflicts are detected and eliminated. 

\subsection{Main Loop}

A train is inserted node by node. For each node, a loop goes through all the possible routes (i.e. combination of incoming, inside and outgoing tracks). The route allowing for the earliest departure time is chosen and one moves on to the next node.

For each route, the arrival and departure times are initialised to their original, unperturbed value or, at departure time set for the previous node plus the minimum time it takes to get to the current node. The constraint checking loop then begins: for each kind of constraint in turn, one checks for violation. If it happens, what we call a \emph{conflict} occurs. This conflict is solved, using either method described below, a flag is raised and the loop is started again.  The loop naturally goes on until no violation flag is raised, meaning $a$ and $d$ are acceptable as such. 

\subsection{Solving Conflicts: moving forward in time}

When a conflict is detected, typically when a train uses a resource
too soon {\bf after} another train, it is trivially solved by moving the troubling variable forward in time until the conflict disappears. For example, if one is currently placing train $c$ in the schedule, one has a rule like :

\begin{eqnarray}
        \exists\ c'\ a(c',i)<a(c,i)<a(c',i)+\alpha\\
        \Rightarrow\ a(c,i):=a(c'i)+\alpha\
\end{eqnarray}

\subsection{Solving Conflicts: kicking obstacles}

It sometimes happens that the previous technique cannot be applied, typically when a train uses a resource too soon {\bf before} another train. Moving forward enough in time would indeed mean taking over the other train on the same track ! In these cases, the technique we use consists in removing the obstacle, i.e. the train creating the conflict, from the schedule. This train is put on the top of the permutation stack and rescheduled right after one is finished with the current train. This usually inverts the conflict and allows for the use of the first conflict solving procedure. Because conflicts can occur anytime during the scheduling process, trains can be removed several times and cycles (a set of trains removing one another) can appear. The number of times a train can be ``kicked'' out of the schedule is therefore limited 

\begin{eqnarray}
        \exists\ c'\ a(c',i)-\alpha<a(c,i)<a(c',i)\\
        \Rightarrow\ remove(c')
\end{eqnarray}

\subsection{Issues}

Two difficulties appear, induced by the ``kicking'' procedure. The first one is that removing previously scheduled trains unstructurates the permutation because the final scheduling order is not the one initially present in the permutation. It means that there is an additional mapping step between an individual's genotype (a sequence of trains) and phenotype (a schedule). This has no practical consequence, but makes the optimisation process harder to understand and therefore to enhance using expert heuristics or problem-oriented metaheuristics.

The second difficulty lies in the fact that not all of the permutations are feasible individuals, i.e., all trains eventually fit in the schedule without violating any constraint. This happens for a small, but non negligible, proportion of the permutations, that can only produce incomplete schedules, leaving a few trains aside because they exceeded the maximum number of ``kicks''. Such permutations are assigned a strongly penalised fitness, and all experiments show that they very quickly disappear, usually as early as after two generations.

\section{The Evolutionary Engine}
\label{evolutionary}
This section gives technical details and parameter settings for the evolutionary part of the system.

\subsection{Representation}

The problem for the Evolutionary Algorithm is to find an optimal
permutation such that, it results in an
optimal schedule when fed into the scheduler.
We chose to use a direct representation: permutations are
straightforwardly encoded using a sequence of integers. 
Other representations were considered, like the Edge Representation
\cite{Whitley_TSP89,Radcliffe-variance}, 
or the Random Keys \cite{randomKeys,RKGA}.
However,  edges are here meaningless, and Random Keys would add
yet another mapping step between genotype and phenotype,
making it even harder to understand how the EA navigates through
the search space. Additionally, 
beyond its simplicity in terms of
implementation, direct representation is here better
suited for the design of specific, intelligent operators based on
expert knowledge of the problem, which is often the key to success
for difficult real-world problems.  

\subsection{Population, selection and replacement scheme}

A standard $(\mu+\lambda)$ replacement scheme is borrowed from Evolution Strategies. It means that we have a population of $\mu$ parents who produce $\lambda$ children using variation operators (see below). Among this new population composed of both parents and offspring, we choose the best $\mu$ individuals as the  new parents for the next generation.

The selection of which parents will reproduce is made using a ``deterministic'' tournament of size $s$ (i.e. the best out of $s$ uniformly chosen parents is returned each time). We have not investigated much of the impact of this choice on the efficiency of the algorithm, and have chosen this particular scheme for it is known to be robust and efficient. We nonetheless have observed that the selection pressure, $s$, has an impact on how quickly infeasible individuals disappear at the beginning of the search. It should be therefore set to a rather high value, without, of course, going too far and causing premature convergence.

\subsection{Variation operators}

Several traditional permutation-based variation operators have been implemented, but extensive experimentation showed that using a simple, although enhanced, swap mutation (see below for details)  was the best choice. Among other variation operators that have been tested are some variant of the Partially Matched Crossover (PMX \cite{Goldberg89}), the Uniform Order Crossover (UOX \cite{Goldberg89}), the half-Swap mutation and a variety of enhancements inspired by metaheuristics (Tabu search, simulated annealing, etc.). For space reasons, only the retained Swap Mutation will be detailed here.

\subsubsection{The Swap Mutation}

This operator simply consists in swapping two elements of the permutation, which means two trains exchanging their ranks in the permutation. To prevent it from being too disruptive, this swapping takes place within a {\bf restricted radius} $R$, which means the two swapped elements cannot be distant of more than $R$ permutation spots. This swapping operation can be repeated $T$ times.

Drawing an analogy from {\bf Simulated Annealing}, we set the tradeoff between exploration and exploitation by controlling the number of times this swap is performed by each mutation operation. This parameter $T$ (for \emph{Temperature}) decreases along with generations according to the following monotonous function:

\begin{equation}
T(n)=
\begin{cases}
T_0 & \text{if $n<n_0$}, \\
T'(n) & \text{otherwise}.
\end{cases}
\end{equation}

\begin{equation}
T'(n)=T_\infty+2(T_0-T_\infty)(1-\frac{1}{1+e^{-\gamma(n-n_0)}})
\end{equation}

Additionally, and following Radcliffe et al. \cite{radcliffeBinMut}, we make the choice of T less deterministic and narrow by replacing its value by a realisation of a {\bf binomial law} centered around this value. This aims at a discrete approximation of a Gaussian mutation, making $T$ the most likely value but neighbouring values possible as well. This is interesting especially in the late stages of the search when $T(n)$ has converged to $T_\infty$ as it allows for both very local and explorative mutations. 

\begin{figure}[htbp]
    \begin{center}
        \epsfig{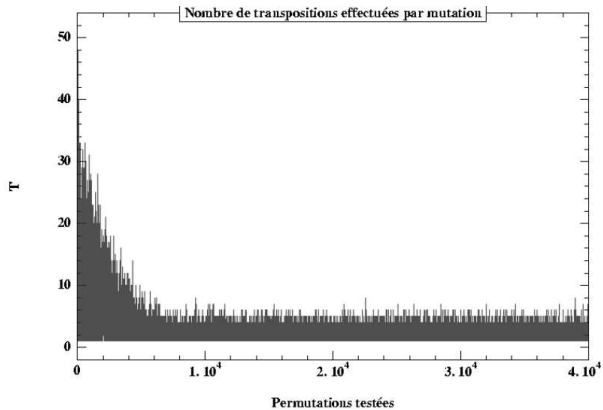}      
      \caption{T varying with time}
      \label{fig:mutation}
    \end{center}
\end{figure}

To help the reader visualise the evolution of T with time, a typical profile is given by Figure \ref{fig:mutation}.

\section{Results}

For the sake of brevity and simplicity, this section only reports results obtained by the final algorithm, after calibration at its empirical best, and on the largest and most difficult problem instance we dealt with. The interested reader can refer to \cite{} for more exhaustive experimental data.

The instance we work with holds around 500 nodes and 541 trains, which yields a total of 1 million variables and 2 millions constraints, all categories included. The perturbation that originates the problem comes from a train being delayed for 10 minutes at a large connecting node around the middle of the considered time period.

Because we are dealing with two very different and therefore hard to compare kinds of algorithms, we chose to use execution time on similar computers as the basis for efficiency comparison. Somewhat relaxing the real-time constraint, we set the maximum computation time to four hours, which empirically appears as a sufficient value to allow our lightly optimised search algorithm to express itself without nearing the pit of potentially infinite computation time. All results are averaged over 11 runs. As illustrated by figure \ref{fig:fourHours}, over these four hours, CPLEX alone manages to go from a fitness of about 50000 to 15000 approximately. Longer runs (24 hours) suggest the global optimum is situated around 12750. A solution is considered good when it reaches 30000 and excellent when it reaches 20000. One can see on the curve that the optimisation is not smooth: long periods of time take place by without any improvement as the Branch-and-Bound ``blindly'' explores the search space; when an improvement occurs however, its amplitude can be very large.

\subsection{Evolutionary Algorithm alone}

The evolutionary algorithm is first applied alone, using the parameters given in table \ref{tab:GAparams}.

    \begin{table}[h]\label{tab:GAparams}
      \begin{center}
        \begin{tabular}{|c|l|}
          \hline
          \multicolumn{2}{c}{\textbf{Parameters}} \\
          \hline
          \hline
          Name & Value \\
          \hline
          $\mu$ & 10\\
          $\lambda$ & 70\\
          $s$ & 2\\
          $n_0$ & 0\\   
          $T_0$ & 50\\
          $T_\infty$ & 1\\
          $R$ & 12\\
          \hline
        \end{tabular}
      \end{center}
      \caption{GA parameters}
    \end{table}

\begin{figure*}[htbp]
    \begin{center}
  \epsfig{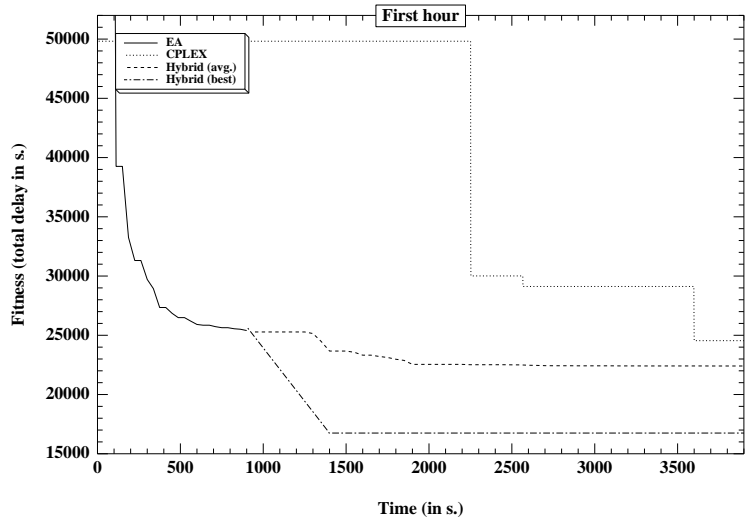}      
      \caption{Results over the first hour}
      \label{fig:firstHour}
    \end{center}
~\\ 
\end{figure*}

As shown in figure \ref{fig:firstHour}, the EA alone only needs 15 minutes (20 generations) to reach, with a high degree of confidence when averaged over several dozen of runs, a fitness region between 25000 and 26000. This is not only much faster than CPLEX alone, but also much more progressive: the optimisation curve is smoothly decreasing whereas that of CPLEX shows staircase-like outlook. Additionally, CPLEX does not even decrease from its initial value during the first half an hour.

Such a smooth optimisation curve might indeed prove very useful in a real-time, interactive context: the algorithm is a real {\em any-time algorithm} where operators are able to stop the optimisation when needed and nevertheless get an improved solution.

Moreover, a strong advantage of evolutionary algorithms is that they are fully parallel: Given adequate computational resources, computation time can be divided by the population size, $\lambda$ in our case. In the present configuration, this would bring the computation time down from 15 minutes to a few seconds, getting close to real-time performance for this problem.

\subsubsection{Sub-optimality}

Beyond these 15 minutes, however, the EA seems  unable to improve the solution ny longer, suggesting the EA search space global optimum has been reached. Though this level of fitness is far from the actual optimum, it seems to be nevertheless acceptable from an expert's viewpoint. Let us discuss in turn two possible reasons for such a sub-optimality.

The first reason comes from the mapping between the genotypes (permutations) and the phenotypes (schedules). There is no formal proof that this mapping is surjective. Schedules can therefore exist that correspond to no permutation whatsoever. If the optimal schedule lies among those non-reachable schedules, the EA will obviously never find it. Both intuition and litterature however suggest that permutations point to the best region of the schedule space and that this kind of indirect approach is a good heuristic.

The second reason comes from the scheduler itself.
As opposed to classical indirect permutation-based approaches, the scheduler was here designed to be semi-greedy instead of actually greedy (see Section \ref{scheduler}). Train insertions are only optimal at the node level, not globally, and this is almost certainly the main cause of sub-optimality. Indeed, detailed examination of (manually designed) successful schedule indicates that an important trick to improve a schedule is to allow trains to overtake each other at key nodes. This implies, in most cases, an artificial delay for one of the trains at the node, while it waits to be overtaken. And our scheduler is unfortunately unable to find these efficient ``building blocks'', inducing undoubtedly a great loss on the way to optimal schedules. The reason why we have not yet improved the scheduler in that direction is twofold. First, making the scheduler less shortsighted would require the implementation of some sort of backtracking or forwardtracking system. This would come at a non negligible price in terms of development and would require a much more complex architecture. Second, the search space would then be much larger and the algorithm all the slower, depriving it from the efficiency which makes it interesting as of now.  

\subsection{Hybrid Algorithm}

\begin{figure*}[htbp]
    \begin{center}
  \epsfig{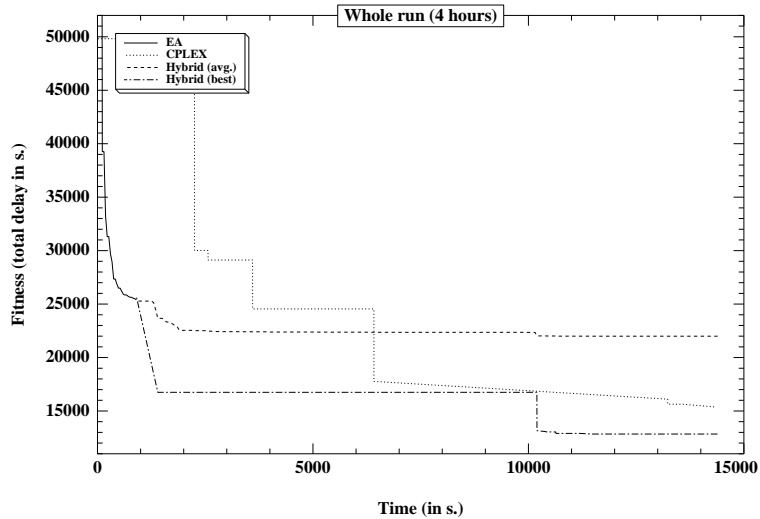}      
      \caption{Results over the whole four hours run}
      \label{fig:fourHours}
    \end{center}
~\\ 
\end{figure*}

The hybridisation of the EA with CPLEX, as described in Section \ref{global}, allows us to overcome this limitation. Once the population has converged, its best individual is fed to CPLEX as a starting point. Figures \ref{fig:firstHour} and \ref{fig:fourHours} show how it goes, over the first hour and over the whole four hours respectively. 
A distinction needs to be done between the best case and the average case. Over the first hour, both are better than CPLEX alone. Looking further in time, CPLEX takes over the hybrid a little before the second hour in the average case. In the best case however, the hybrid is both much faster (it reaches a solution around 16000 in 25 minutes, vs more than 3 hours for CPLEX) and more efficient (it reaches better solutions, around 12800, at the end of the run vs 15000 for CPLEX alone). 

Analysis of the various runs actually show that they are divided in two categories with low internal variance: the runs from the first category hardly improve over the four hours, moving slowly from 25000 to 23000; the runs from the  other category, among which the best case shown discussed above, reach a fitness level around 16000 very quickly, after only a few minutes. The ratio between these two categories is about of 1 to 4 in favour of the worst case. Because all these runs start from similar fitness values (slightly above 25000), it seems there might very important differences between two excellent solutions that cannot be signalled to the EA. These differences obviously matter a lot to CPLEX and greatly affect how well its Branch and Bound search algorithm is going to perform on those solutions. The nature of these differences remains to be investigated at this stage of the work.

\section{Conclusion}

Our algorithm, whether alone or hybridised, obtains convincing results on a
large, complex instance of a difficult scheduling problem. It is
extremely efficient in the early stage of the search and quickly reaches
a satisfactory zone of excellence. Thanks to its parallelisation
potential, our algorithm might even open unexpectedly soon the door
to real-time exploitation. additionally, over long runs, it has shown
its ability, in a non negligible minority of cases, to perform much
better than CPLEX alone in terms of both speed and quality of the final
solution. 

There are, however, two important drawbacks to our approach. The first
is that much of the efficiency of an indirect approach relies on the
mapping function whose design and implementation are costly and demand
a great deal of expert knowledge on the problem and data. Such requisites
consumed the greater part of our time and efforts. This fact somewhat
weakens, for such problems, the view of Evolutionary Algorithms as
generic, easily adaptable optimisation techniques. However, it does not
affect the efficiency of the finally obtained algorithm. The
second drawback is in the sub-optimality of the best solutions obtained
by the evolutionary algorithm, which at this stage prevent it from
being used alone.   

Further work on the genetic side will focus on reducing sub-optimality
without loosing too much on computation time and on the enhancement of
the evolutionary engine and operators using metaheuristics based on
elementary problem-related information, such as individual delay or
distance to the perturbation.  

\section*{Acknowledgements}

\bibliographystyle{plain}
\bibliography{semetCEC05}

\end{document}